# Enriching Consumer Health Vocabulary Using Enhanced GloVe Word Embedding

Mohammed Ibrahim [0000-0001-6842-3745], Susan Gauch [0000-0001-5538-7343], Omar Salman [0000-0003-4797-6927], and Mohammed Alqahatani [0000-0002-8872-6513]

University of Arkansas, Fayetteville AR 72701, USA
{ msibrahi,sgauch,oasalman,ma063}@uark.edu

**Abstract.** Open-Access and Collaborative Consumer Health Vocabulary (OAC CHV, or CHV for short), is a collection of medical terms written in plain English. It provides a list of simple, easy, and clear terms that laymen prefer to use rather than an equivalent professional medical term. The National Library of Medicine (NLM) has integrated and mapped the CHV terms to their Unified Medical Language System (UMLS). These CHV terms mapped to 56000 professional concepts on the UMLS. We found that about 48% of these laymen's terms are still jargon and matched with the professional terms on the UMLS. In this paper, we present an enhanced word embedding technique that generates new CHV terms from a consumer-generated text. We downloaded our corpus from a healthcare social media and evaluated our new method based on iterative feedback to word embeddings using ground truth built from the existing CHV terms. Our feedback algorithm outperformed unmodified GLoVe and new CHV terms have been detected.

## 1 Introduction

With the advancement of medical technology and the emergence of Internet social media, people are more connected than before. Currently, many healthcare social media platforms provide online consultations for patients. The Pew Research Center reported that in 2011 about 66% of Internet users looked for advice regarding their health issues[1]. Furthermore, the rate of using social media by physicians reached to 90% in 2011 [2]. Physicians, nurses, or any expert who practices medicine will not be able to interact effectively with laymen unless they have a lexical source or ontology that defines all medical jargon in an easy and clear way to be understood by laymen.

A concentrated effort, involving experts in different health fields, has created several medical ontologies. These ontologies, such as Mesh, SNOMED CT, and many others, were built to describe and connect professional medical concepts. The United States National Library of Medicine (NLM) combined many resources into one thesaurus called the Unified Medical Language System (UMLS). The UMLS is a metathesaurus consisting of more than 3,800,000 professional biomedicine concepts [3].

In contrast to the UMLS, the Open-Access and Collaborative Consumer Health Vocabulary (OAC CHV, or just CHV for short), is a collection of medical terms written in plain English. It provides a list of easy terms to refer to a professional medical concept. The goal of developing CHV terms is to lessen the gap between laymen and medical experts and to improve the accuracy of health information retrieval [4]. Out of 3,800,000 concepts on the UMLS, only 56,000 concepts assigned a CHV term(s).

In spite of the claim that the CHV contains laymen terms mapped to professional concepts, our investigations showed that out of the 56,000 CHV terms assigned to UMLS concepts, 27,000 (48%) of concept's terms are still jargon and are just morphological variations of the professional term. For these, the CHV terms contain either downcased letters, the plural 's', or numbers and punctuations.

To address this, we propose a system that processes consumer-generated text from a healthcare platform to find new CHV terms. The system uses the Global vectors for word representations (GloVe). We also improved this algorithm by applying an automatic, iterative feedback approach.

## 2   Related Work

Building a lexical resource or an ontology with the help of human can lead to a precise, coherence, and reliable knowledge base. However, it involves a lot of human effort and consumes a lot of time. To address that, Kietz et al. [5] prototyped an approach to build a company ontology semi-automatically using an existing ontology and a company-related dictionary. Harris and Treitler [4] developed the Open Access Consumer Health Vocabulary (CHV) using statistical approaches and text collected from the internet. There are several methods have been proposed to enrich this consumer vocabulary, such as He et al.[6] proposed by enriching the CHV vocabulary using a similarity-based technique. They collected posts from a healthcare social media and applied the k-means algorithm to find similar terms. However, their work is tied to drawbacks of the K-means algorithm, such as the number of clusters and how to initialize these clusters. Gu [7] also tried to enrich CHV vocabulary by applying three recent word embedding methods. However, his work is not completely automatic and involves human. In contrast, our proposed system is completely automatic and uses state-of-the-art methods to extract new CHV terms.

## 3   Methodology

To enrich the CHV terms, we need a corpus that contains many laymen terms for medical concepts. Medhelp.org is a healthcare social media platform in which people post information about their health issues. These posts are presented in a question/answer format wherein people share their experiences, knowledge, and opinions within different health communities [8]. Such healthcare social media can be an excellent source from which to extract new CHV terms.

The other requirement for our system is a set of medical concepts that have associated laymen terms to use as ground truth. For that, we used the already existing CHV vocabulary. For each concept, we selected one term as the seed and judge the algorithms by their ability to detect synonyms, i.e., the other terms in the concept. We use the seed terms to locate contexts in corpus from which the new CHV vocabulary might be extracted. Figure 1 shows the steps of our system.

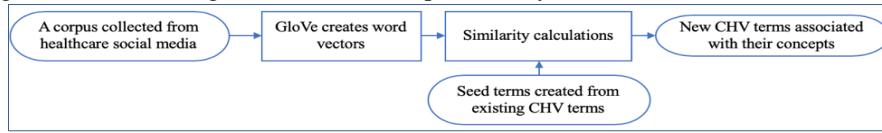

**Fig. 1.** The methodology of extracting new CHV terms

### 3.1 GloVe Embeddings

GloVe algorithm is a word vector representation method. It builds word embeddings using a log bilinear model. This algorithm combines the advantages of local window methods and global matrix factorization [9]. GloVe has many hyperparameters that can affect its results, but the window size and vector dimension are the most effective parameters. This paper reports the results using the same setting reported in [9].

### 3.2 GloVe Iterative Feedback (GloVeIF)

The idea here is feeding the most similar terms that the GloVe algorithm produces again to the GloVe cooccurrence matrix. This method explores the potential source of auxiliary information, the corpus itself, through a process of iterative feedback. Figure 2 shows the steps of this method. We have highlighted the iterative feedback steps with orange. In this method, GloVe lists the most similar terms to the CHV terms that iteratively fed back to GloVe to boost the frequencies in the co-occurrence matrix as though there were additional contexts available. When GloVe trains its word vectors, a seed term for every medical concept will be chosen, and a list of top $n$ most similar terms will be listed. Our GloVeIF algorithm then iteratively submits these top $n$ terms to the trained vectors to find their top $k$ most similar terms. Then, it adds them to the top $n$ list. For example, if top $n = 10$ and top $k = 5$, then the final list of most similar terms for every seed term is going to be (10*5)+10 = 60 terms. After having this list ready, GloVeIF feeds this list back to the GloVe model. So, we have two feedbacks. The first from the GloVe pre-trained vectors to expand top n similar terms, and the second from the GloVe model.

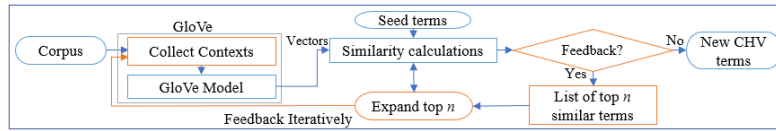

**Fig. 2.** The GloVeIF architecture

## 4 Evaluation

**Corpus and Seed Terms.** We build our laymen corpus from different MedHelp communities. We downloaded all the questions on these communities on April 20 of 2019. The dataset size is roughly 1.3 Gb and contains approximately 135,000,000 tokens. The corpus cleaned from punctuations, numbers, and traditional stopwords. We removed any word with length less than 3. Also, we created our special stopword list to remove common medical terms such as 'test', 'procedure', and 'disease'. We did the same process to our seed term list. Moreover, the seed terms list is cleaned to remove words that are morphological variants of professional medical concepts within the UMLS, so we kept only true laymen terms. Finally, only concepts that have at least two associated CHV terms that occurs 100 times are kept in the seed term list so that we will have enough contexts within the corpus to attempt word embeddings. The final list contains 1257 concepts along with their associated CHV terms. Table 1 shows some of the UMLS professional concepts and their associated CHV terms.

Table 1. Example of UMLS concepts with their associate CHV terms.

| CUI | Medical Concept | CHV terms | | | |
|---|---|---|---|---|---|
| C0035334 | retinitis pigmentosa | pigmentary | retinopathy | cone | rod |
| C0034194 | pyloric stenosis | stenos | gastric | outlet | obstruct |

**Baselines and Evaluation Metrics.** We compared the GloVeIF with the baselines as reported in [7]. One of the baselines is the GloVe itself, and the other two are the Word2Vec[10] and FastText[11]. For the ground truth dataset, we used the seed term list created from the existing CHV vocabulary. A random term picked from the concept's associated CHV terms to be used as a seed term. The top 10 most similar terms to that term are listed and compared with the other left CHV terms. For our accuracy measurements, the precision, recall, F-score, and mean reciprocal rank (MRR) applied to measure the performance of the system. Also, we measured the average of the number of concepts that the algorithm was able to detect.

## 5  Results and Discussion

Table 2 shows the results of implementing the GloVeIF and the comparisons with the baselines. All the algorithms run with their basic settings except the vector dimension is set to 100 and window size of ±10. All the results reported with top $n = 10$, and for the GloVeIF, the top $k$ is set to 5.

Table 2. Results of running two algorithms.

| Algorithm | Precision (%) | Recall (%) | F-score (%) | MRR | Concepts |
|---|---|---|---|---|---|
| **Word2Vec** | 21.24 | 16.66 | 18.61 | 0.26 | 9 |
| **GloVe** | 15.86 | 12.5 | 13.98 | 0.35 | 18 |
| **GloVeIF** | 17.56 | 13.39 | 15.19 | 0.27 | 21 |

Our GloVeIF algorithm outperformed the basic GloVe algorithm with an 8.7% improvement in the F-score. GloVeIF also outperformed all other algorithms by identifying the highest number of concepts. However, the basic GloVe algorithm was the best in terms MRR. On average, the basic GloVe and GloVeIF, found the true synonym to the seed term in the 4[th] position of ranked similar terms. Table 2 shows that the Word2Vec algorithm with a continuous bag of words (CBOW) model outperformed other algorithms in term of the precision, recall, and F-score. However, it was only able to find synonyms in half or fewer of the concepts versus the others. We did not report the results from the FastText algorithm and the Word2Vec (Skip-Gram model) because they were unable to detect any CHV terms related to the seeds at all.

The ground truth list has about 1200 concepts along with their associated CHV terms. Among the 5 algorithms tested, our GloVeIF algorithm was able to detect associated terms for the largest number of concepts. However, at 21, this was still very low. We believe that the number of concepts was so low because the evaluation used the existing CHV terminology. We previously mentioned that 48% of the existing CHV terms were just morphological variations of the UMLS terms (and thus removed from our data set). However, many of the remaining CHV terms are not truly laymen's terms and the users tend to be laymen asking questions and use common terms only, or experts providing answers who use professional terms only, so the seeds and the candidate CHV terms do not do co-occur frequently enough to be discovered by word embedding methods. However, the raw results of GLoVeIF are quite promising; it seems to do an excellent job identifying laymen's terms that are not in the current CHV. We displayed a list of 500 seed terms, along with their topmost similar term to three judges[1] to rate their relatedness as 1 (related) or 0 (not related). This informal, human validation found that 80% of time the most similar term was related seed term. Table 3 shows some of the results that the GloVeIF detected. The most similar terms in Table 3 sorted by their degree of similarity to the seed terms.

**Table 3.** Some of the seed terms and their most similar terms that GloVeIF produced.

| Seed term | Term1 | Term 2 | Term 3 | Term 4 |
|---|---|---|---|---|
| **bowel** | bladder | constipation | diarrhea | intestine |
| **skin** | itch | itchy | dry | irritate |
| **ray** | xray | scan | mri | spine |

Since MedHelp.org posts are all related to health issues, we can see that some of the seed terms are general, such as the term *skin*, but their most similar terms are health issues such as *skin itching*, *dry skin*, and *skin-irritating*. Although clearly related to the seed term, none of the candidate terms were in the associated CHV concept.

---

[1]. The judges were the first, third, and fourth authors of this paper.

## 6 Conclusion and Future Work

This paper presents a method to enrich a consumer health vocabulary (CHV) with new terms from a healthcare social media texts using word embeddings. We begin by demonstrating that the CHV contains many terms that are not true laymen's terms. Our algorithm, GLoVeIF, is automatic, and identifies new terms by locating synonyms to seed CHV terms. We conclude by demonstrating that many of the top synonyms proposed by GLoVeIF were related terms, even though they do not appear in the current CHV. For future work, we suggest more investigation regarding the number of concepts and their CHV terms that can be detected. We also suggest implementing the same seed term approach using more recent word embedding methods.

## References

1. S. Fox, "Health Topics," *Pew Research Center: Internet, Science & Tech*, 01-Feb-2011. https://www.pewinternet.org/2011/02/01/health-topics-3/ (Oct. 21, 2019).
2. M. Modahl, L. Tompsett, and T. Moorhead, "Doctors, Patients & Social Media," *Social Media*, p. 16, 2011.
3. O. Bodenreider, "The unified medical language system (UMLS): integrating biomedical terminology," *Nucleic acids research*, vol. 32, no. suppl_1, pp.
4. K. M. Doing-Harris and Q. Zeng-Treitler, "Computer-assisted update of a consumer health vocabulary through mining of social network data," *Journal of medical Internet research*, vol. 13, no. 2, p. e37, 2011.
5. J.-U. Kietz, A. Maedche, and R. Volz, "A Method for Semi-Automatic Ontology Acquisition from a Corporate Intranet," p. 15, Oct. 2000.
6. Z. He, Z. Chen, S. Oh, J. Hou, and J. Bian, "Enriching consumer health vocabulary through mining a social Q&A site: A similarity-based approach," *Journal of biomedical informatics*, vol. 69, pp. 75–85, 2017.
7. G. Gu *et al.*, "Development of a Consumer Health Vocabulary by Mining Health Forum Texts Based on Word Embedding: Semiautomatic Approach," *JMIR Medical Informatics*, vol. 7, no. 2, p. e12704, 2019, doi: 10.2196/12704.
8. H. Kilicoglu *et al.*, "Semantic annotation of consumer health questions," *BMC bioinformatics*, vol. 19, no. 1, p. 34, 2018.
9. J. Pennington, R. Socher, and C. Manning, "Glove: Global vectors for word representation," in *Proceedings of the 2014 conference on empirical methods in natural language processing (EMNLP)*, 2014, pp. 1532–1543.
10. T. Mikolov, I. Sutskever, K. Chen, G. S. Corrado, and J. Dean, "Distributed representations of words and phrases and their compositionality," in *Advances in neural information processing systems*, 2013, pp. 3111–3119.
11. P. Bojanowski, E. Grave, A. Joulin, and T. Mikolov, "Enriching word vectors with subword information," *Transactions of the Association for Computational Linguistics*, vol. 5, pp. 135–146, 2017.